\documentclass[runningheads,a4paper]{llncs}

\usepackage{amssymb}
\usepackage{graphicx}

\usepackage{url}
\urldef{\mailsa}\path|{nicolas.riche,matthieu.duvinage, matei.mancas,|
\urldef{\mailsc}\path|bernard.gosselin,thierry.dutoit}@umons.ac.be|    
\newcommand{\keywords}[1]{\par\addvspace\baselineskip
\noindent\keywordname\enspace\ignorespaces#1}

\begin{document}

\mainmatter  
\title{A study of parameters affecting\\ visual saliency assessment}
\titlerunning{Lecture Notes in Computer Science: Authors' Instructions}

\author{Nicolas Riche\footnotemark[1]  \and Matthieu Duvinage\footnotemark[1] \and Matei Mancas  \and \\ Bernard Gosselin\and Thierry Dutoit }
\footnotetext[1]{N. Riche and M. Duvinage contributed equally to this work}
\authorrunning{Lecture Notes in Computer Science style}

\institute{  University of Mons (UMONS), Faculty of Engineering (FPMs),\\ 20, Place du Parc, 7000 Mons, Belgium \\ \mailsa\\ \mailsc\\}
\toctitle{Lecture Notes in Computer Science}
\tocauthor{Authors' Instructions}
\maketitle

\begin{abstract}
Since the early 2000s, computational visual saliency has been a very active research area. Each year, more and more new models are published in the main computer vision conferences. Nowadays, one of the big challenges is to find a way to fairly evaluate all of these models. In this paper, a new framework is proposed to assess models of visual saliency. This evaluation is divided into three experiments leading to the proposition of a new evaluation framework. Each experiment is based on a basic question: 1) there are two ground truths for saliency evaluation: what are the differences between eye fixations and manually segmented salient regions?, 2) the properties of the salient regions: for example, do large, medium and small salient regions present different difficulties for saliency models? and 3) the metrics used to assess saliency models: what advantages would there be to mix them with PCA? Statistical analysis is used here to answer each of these three questions.

\end{abstract}

\section{Introduction}

The aim of saliency algorithms is to automatically predict human attention and to select a subset of all the perceived stimuli. Bottom-up attention uses signal-based features to find the most salient objects. Top-down attention uses \textit{a priori} knowledge about the scene or task-oriented knowledge to modify the bottom-up saliency. In this paper, saliency algorithms are only based on the bottom-up approach. \\

The literature is very active concerning saliency algorithms due to their wide potential applications including image compression, object detection, scene classification, robotics, \textit{etc}. 
Due to the explotion of studied models, some evaluation studies (such as \cite{rare-journal}, \cite{vikram}, \cite{frintrop11} and \cite{toets}) and online benchmarks (like \cite{judd12} and  \cite{ali1}) have been proposed.
A key underlying issue is: {\bfseries how to fairly evaluate all of these models?} For the specific field of robotics, for example, saliency models should be able to cope with real-life stimuli where regions of interest have different sizes. The algorithms purpose is to model human eye fixations, but also to detect and segment objects of interest. In this paper, we investigate three important aspects of visual saliency assessment in real-life images: 1) which ground truths should they use?, 2) what is the relation between the models performance and the size of the salient regions into images? and 3) what advantages would there be to mix comparison metrics ?  

First of all, there are mainly two ground truth categories to assess a saliency map: 1) human eye fixations obtained using an eye tracker device and 2) manually segmented and labelled salient regions. In our study, we analyse the difference and the coherence between them. The second aspect of this study is about different categories of salient regions. Are saliency models equally efficient in predicting human gaze on three categories of salient regions: large, intermediate and small? This is an important issue as real-life objects and scenes contain very wide range of objects sizes. Finally, various evaluation measures exist to compare saliency and ground truth maps. In the third experiment, we use the first component of PCA (maximum variability) for a given ground truth to evaluate saliency models. The results of using a mix of metrics are compared with the use of metrics alone. \\ 

In section \ref{sec:methods}, the methods used in the three experiments (database, saliency models and comparison metrics) are defined. In section \ref{sec:expe}, we detail the three experiments leading us to a comparison between 12 state-of-the-art saliency models. Finally, we conclude in section \ref{sec:conclusion}. 

\section{Method}
\label{sec:methods}
\subsection{Database}

\begin{figure}[h!]
 \centering
 \includegraphics[scale=0.22]{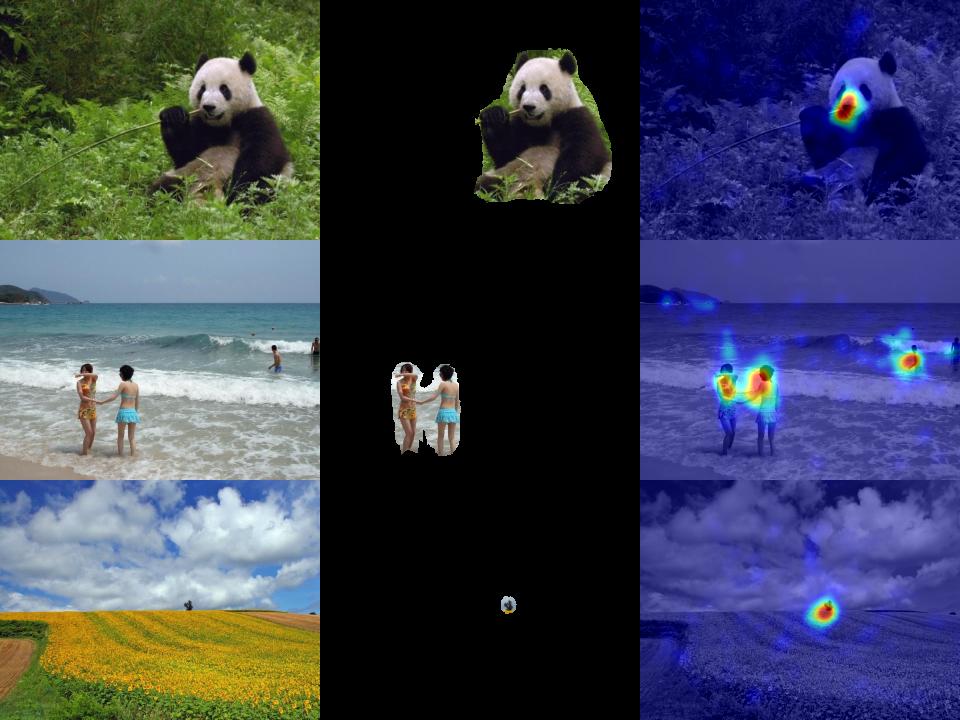}
 \caption{Sample images of the Jian Li's database using three sizes of regions of interest. By rows: large, intermediate and small salient regions.}
 \label{fig:base}
\end{figure}

The database used here has been published by Jian Li \cite{li} and provides both region ground truth (human labelled) and eye fixation ground truth (collected with an eye tracker). 
The whole database can be divided into six different groups of images, which constitutes another interesting property of this dataset. Among these six groups three of them are considered for our study: large, intermediate and small salient regions as depicted in Fig. \ref{fig:base}. This data set contains 235 color images. The detailed properties of the dataset (how small, medium and large regions are chosen, number and characteristics of the the observers, manual labelling protocol, ...) can be found in \cite{li}.

\subsection{Bottom-up saliency models}

A lot of saliency models based on different approaches exist and therefore, it's hard to find taxonomy to classify them. In this study, we decide to use Ali Borji's taxonomy \cite{ali}, in which models are sorted based on their mechanism to obtain saliency maps. Twelve state-of-the-art models are assessed in this study. They represent a wide range of saliency approaches detected by Ali Borji. In case of some models fall into more than one category, the classification is made on the primary considered approach.\\

Itti's model \cite{itti} represents the cognitive approach. This model uses three features channels (color, intensity and orientations). From these, center-surround features maps are build and linearly combined to generate the saliency map. SUN \cite{sun} and Torralba \cite{tor} are Baysian models. They propose to define saliency with a Baysian Framework. SUN's model uses natural statistic, by considering what the visual system is trying to optimize when directing attention. Torralba's model uses local and global features. AIM \cite{aim}, DVA \cite{pqft} and RARE \cite{rare} are into the information theoretic category. AIM uses Shannon's self information measure to transform the image feature plane into a visual saliency map. DVA's model introduced the Incremental Coding Length (ICL) approach to measure the respective entropy gain of each feature. RARE's model is based on the idea that locally contrasted and
globally rare features are salient. SR \cite{hz} , PFT \cite{pft}, PQFT \cite{pqft} and Achanta  \cite{achanta} use a spectral analysis approach to compute their saliency map. SR's model develops the spectral residual saliency model based on the idea that similarities imply redundancies. PFT's and PQFT's models are based on the same idea and also proposed to combine more features and use the quaternion Fourier Transform. Achanta implements a frequency-tuned approach to salient region detection using low-level features of color and luminance. Finally, we also decide to used  two other models: HFT \cite{li} and AWS \cite{aws}. AWS predicts very well human gaze into different databases with various metrics. HFT is also an efficient model which is proposed by the database author Jian Li. Moreover, HFT's model is a new saliency detection model by combining global information from frequency domain analysis and local information from spatial domain analysis. AWS's model introduced the Adaptive Whitening Saliency (AWS) model by adopting the variability in local energy as a measure of saliency estimation.

\subsection{Similarity measures}

 A lot of similarity measures exist for saliency models assessment as listed in \cite{lemeur}. In our framework, two metrics have been chosen: NSS and AUROC for their conceptual and statistical complementarity \textbf{\cite{rare-journal}}. Indeed, contrary to the NSS measure, which compares values or amplitudes of the maps, AUROC mainly measures the fixations locations. The codes used in this paper for NSS computation are the freely available Matlab implementation of Borji \cite{nss} and for AUROC, Jian Li's implementation is used \cite{li}. They are applied to both eye-tracking and region-based ground truth. NSS and AUROC measures used here are detailed in the following subsections. 

\subsubsection{Normalized Scanpath Saliency (NSS)}
The Normalized Scanpath Saliency metric was introduced in 2005 by Peeters and Itti \cite{peters}. The idea is to quantify the saliency map values at the eye fixation locations and to normalize it with the saliency map variance:
\begin{equation}
NSS (p) = \frac{SM (p) - \mu_{SM}}{\sigma_{SM}} 
\label{NSS:calc}
\end{equation}
where $ p $ is the location of one fixation and $ SM $ is the saliency map which is normalized to have a zero mean and unit standard deviation. Indeed, the NSS score should be decreased if the saliency map variance is important or if all values are globally similar (small difference between fixation values and mean) because it shows that the saliency model will not be very predictive, 
while he will precisely point a direction of interest if the variance is small or the difference between fixation values and mean high.\\
The NSS score is the average of $ NSS(p)$ for all fixations:
\begin{equation}
NSS = \frac{1}{N} *\sum_{p=1}^{N} NSS(p)
\label{NSS:end}
\end{equation}
where N is the total number of eye fixations.

 \subsubsection{Area under the Receiver Operating Characteristic curve (AUROC)}
Location-based metrics are very popular to evaluate how well a saliency map approximates a human fixation map. They are based on the notion of AUROC coming from signal detection theory. We used here Jian Li's implementation. First, fixations pixels were counted once and the same number of random pixels are extracted from the saliency map. For one given threshold, saliency pixels can be treated as a classifier, with all points above threshold indicated as �fixation� and all points below threshold as �background� For any particular value of the threshold, there is some fraction of the actual fixation points which are labelled as True Positives (TP), and some fraction of points which were not fixation but labelled as False Positive (FP). This operation is repeated one hundred times. Then the ROC curve can be drawn and the Area Under the Curve (AUC) computed. An ideal score is one while random classiffication provides 0.5.\\

To do a fair comparison with both NSS and AUROC metrics, common processing is applied to all models. Indeed, some processing effects will dramatically influence their score. Two well-known issues \cite{li} are the centre-bias and border effect. Centre-bias means that a lot of fixations from natural images databases are located near the image centre because when taking pictures, the amateur photographer often places salient objects in the image centre. The computational saliency models which include a centered Gaussian use the prior knowledge of working on natural images which is not a general assumption, and therefore might be considered as top-down information. Zhang and al. \cite{sun} showed that scores are also corrupted by border effects. If we remove saliency maps borders, scores increase as well because human eye fixations are rarely near the borders of test images. 
Based on \cite{li}, we decide to eliminate the models with an important centre-bias from our framework and apply the same blurring and border cuts for all models to be of equal size and avoids in that way to artificially increase the AUC scores for the models which already do this pre-processing in comparisons with those which do not. The border cut post-processing affecting the fairness during the assessment is thus eliminated.

\section{Experimental Framework}
\label{sec:expe}


\subsection{Experiment 1: Effects of regions or eye-fixations ground-truth}

\begin{figure}[h!]
 \centering
 \includegraphics[scale=0.25]{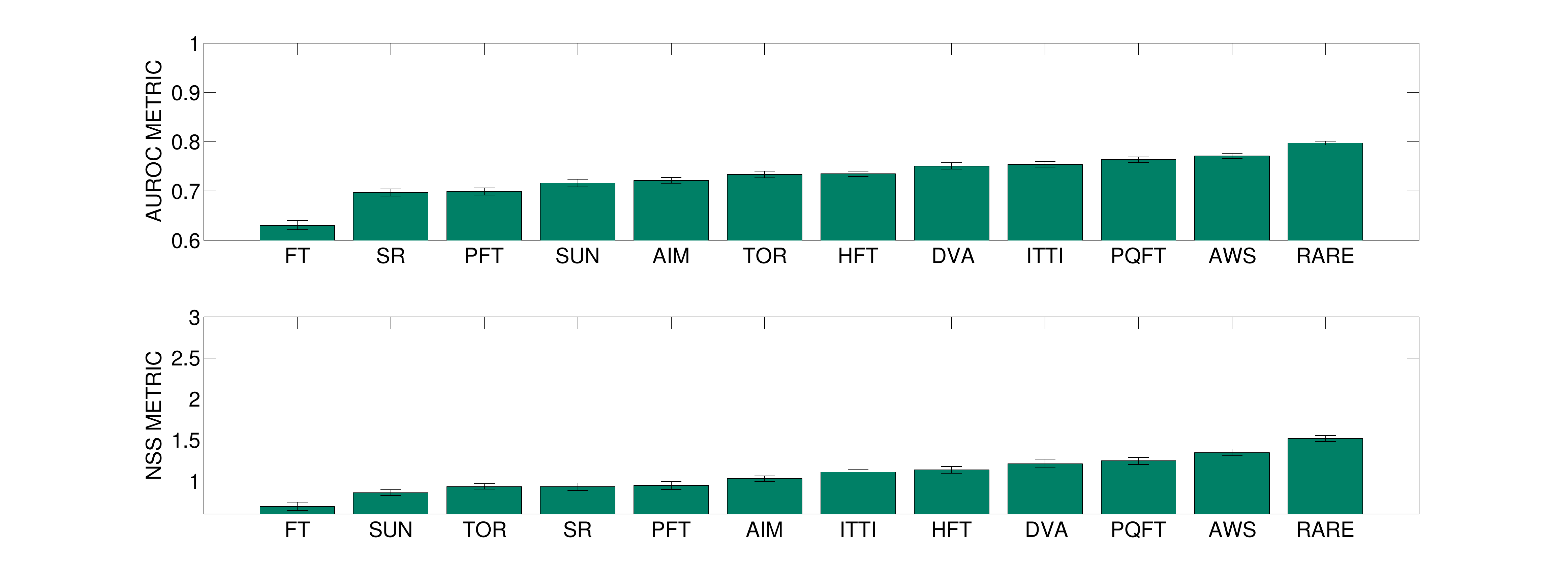}
 \caption{Eye fixations mean scores for all the models on the whole database with their standard deviations}
 \label{fig:fixre}
\end{figure}

\begin{figure}[h!]
 \centering
 \includegraphics[scale=0.25]{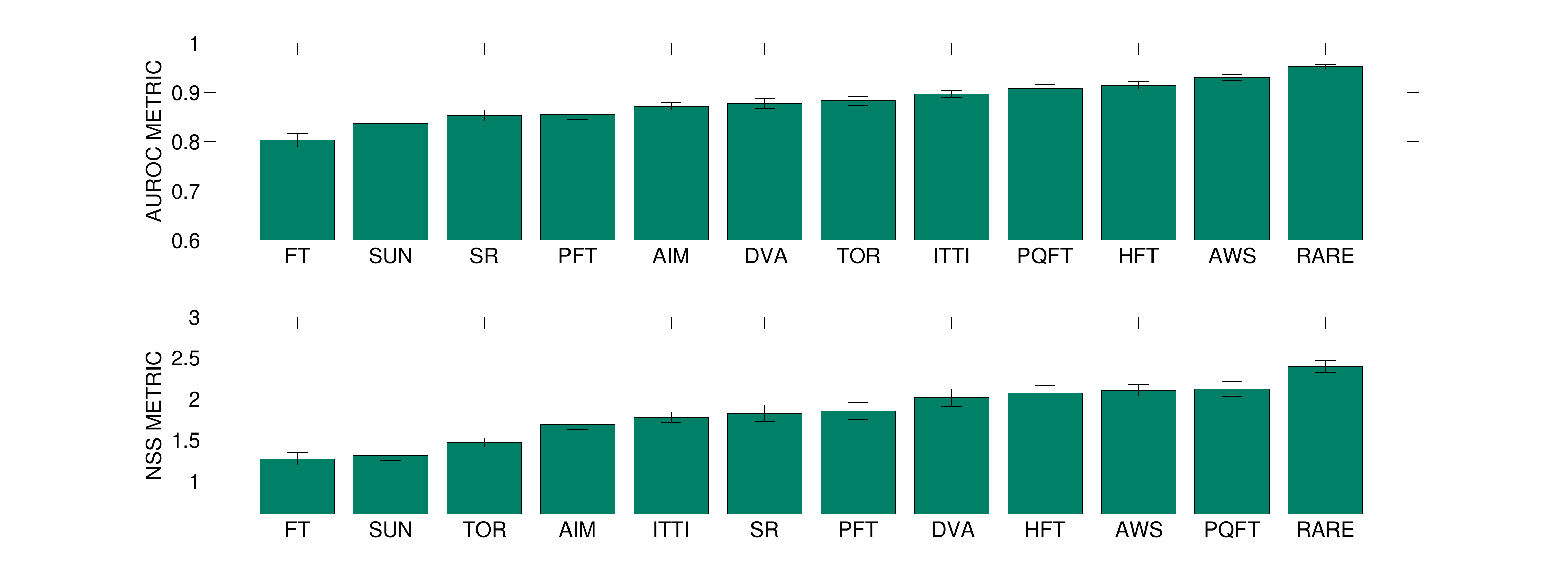}
 \caption{Labelled regions mean scores for all the models on the whole database with their standard deviations}
 \label{fig:zone}
\end{figure}

There are a lot of databases providing either eye-tracking fixations or labelled objects. Jian Li's database has the interest to provide both approaches for the same set of images. Some saliency models will better model eye-fixations while others focus on objects detection and segmentation and are assessed with region-based labelled objects. The main idea of this first experiment is to assess the coherence between the region-based or eye fixation-based ground truths. The mean results of both metrics for each model are computed in Fig. \ref{fig:fixre} and Fig. \ref{fig:zone} for the entire database.
The range of score for both metrics is higher with labelled regions than eye fixations as the latter are much more precise especially for the large salient regions class.

After this first score computation, a ranking-based statistical test is required. Considering our repeated measure design, the two-way Friedman test allows to respond to the $H_{0}$ hypothesis: Are the ranking of the individual results provided by the several models coherent between both ground truth performance evaluation? In our specific design, a blocking operation was performed on images and models to eliminate uninteresting variability and to increase the statistical power (5 \% significant threshold was used).\\

However, significance is not enough and effect-size is at least as important \cite{univarie}. Indeed, significativity only assesses if there is enough evidence to determine whether there is a likely effect between two or more groups. It does not provide information about the size of this effect. The effect size measure by how much the detected effect is significant in practice; in other words, it defines how big the discordance between the region-based and eye fixation-based ground truths is in their image/model evaluation. Unfortunately, there is no specific effect size measure in case of the Friedman test that was adapted to our study. Therefore, we propose to use the Kendall's $W$ concordance measure, which basically fulfils our needs. 

Kendall's $W$ concordance measure  \cite{univarie} is defined in Eq. \ref{Exp:kendall}: 

\begin{equation}
W = \frac{12*S}{m^{2}*(n^{2}n)} 
\label{Exp:kendall}
\end{equation}

where $n$ is the number of models and $m$, the number of metrics. So here $n=12$ $m=2$. $S$, the sum of squared deviations, is defined as in Eq. \ref{Exp:kendall1}:
 
\begin{equation}
S =  \sum_{i=1}^{n} (R_i - \bar{R})^{2}.
\label{Exp:kendall1}
\end{equation} 
 
where $R_i$ is  the total rank given to model i and $\bar{R}$ the mean value of these total ranks.  
 
Kendall's $W$ concordance is a coefficient measuring the degree of agreement between metrics. The value ranges from 0 (no agreement between model ranks) to 1 (full agreement, same models ranking). Furthermore, some rules of thumb are provided to allow the researcher to interpret this measure as depicted in Tab. \ref{W} \cite{univarie}. 

\begin{table}[htbp]
  \centering
  \caption{Interpretation of Kendall's W coefficient}

\begin{tabular}{|c|c|c|}
  \hline
    Kendall's W     &Interpretation & Confidence in Ranks \\

  \hline
 0.5 & Moderate agreement &  Fair \\
 0.7 & Strong agreement    & High \\
  0.9 & Unusually strong agreement    & Very High \\
   1 & Complete agreement    & Very High \\
  \hline
\end{tabular}
  \label{W}
\end{table}

\begin{table}[htbp]
  \centering
  \caption{Concordance based on Friedman test and Kendall Coefficient between fixations and regions results}

\begin{tabular}{|c|c|c|c|}
  \hline
                  & \multicolumn{2}{c|}{Friedman test} & Kendall's concordance \\
                  & p-value  & $\chi^{2}$ (1, N=12) & W \\
  \hline
 AUROC  & 0 &  1765  & 0.82 \\
 NSS        & 0  &  688.25  & 0.88 \\
  \hline
\end{tabular}
  \label{conc}
\end{table}

As shown in Tab. \ref{conc} (Friedman test), although differences between eye fixations and regions results are significant, the Kendall's concordance between both ground-truths is very good for each metric. 
This means that there is a difference between both rankings, but the size of this difference based on Kendall's W coefficient is relatively small  \cite{rare-journal}.
 If models have good results with one ground-truth, it is quite unlikely that these models completely fail with the other ground-truth except due to statistical fluctuation. A saliency model which is good in predicting human eye fixations will remain good in predicted human labelled regions and conversely.  
However, we decide to work with both of ground truth for the next experiment because of significant differences.


\subsection{Experiment 2: Effects of Large, Medium and Small regions}

In this experiment, we want to compare the effectiveness of the models on three different images categories (large, medium and small salient regions). In real-life images, all kinds of objects sizes can be seen and saliency models which are tuned for a given object size are not suitable.
 
\begin{figure}[h!]
 \centering
 \includegraphics[scale=0.28]{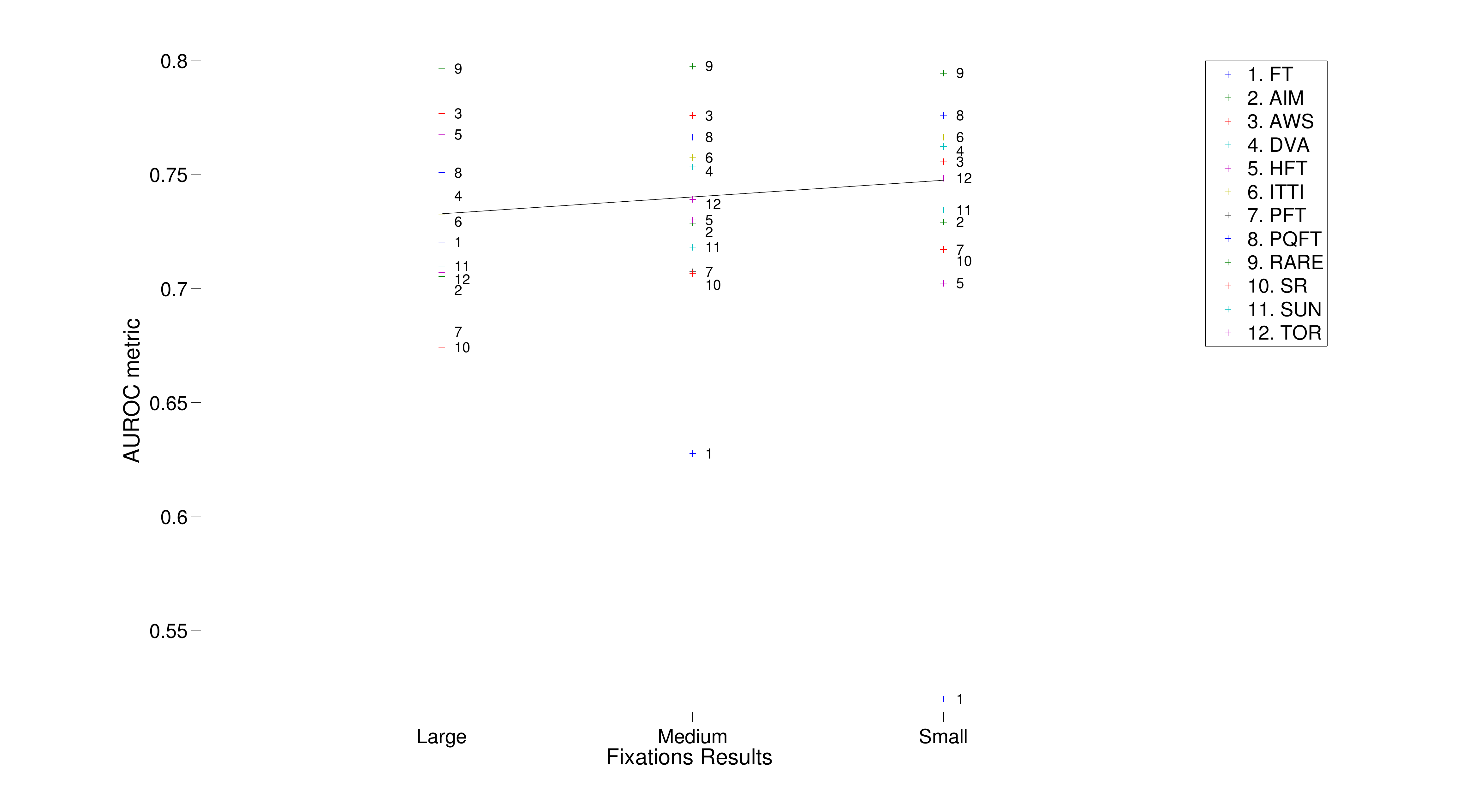}
  \includegraphics[scale=0.28]{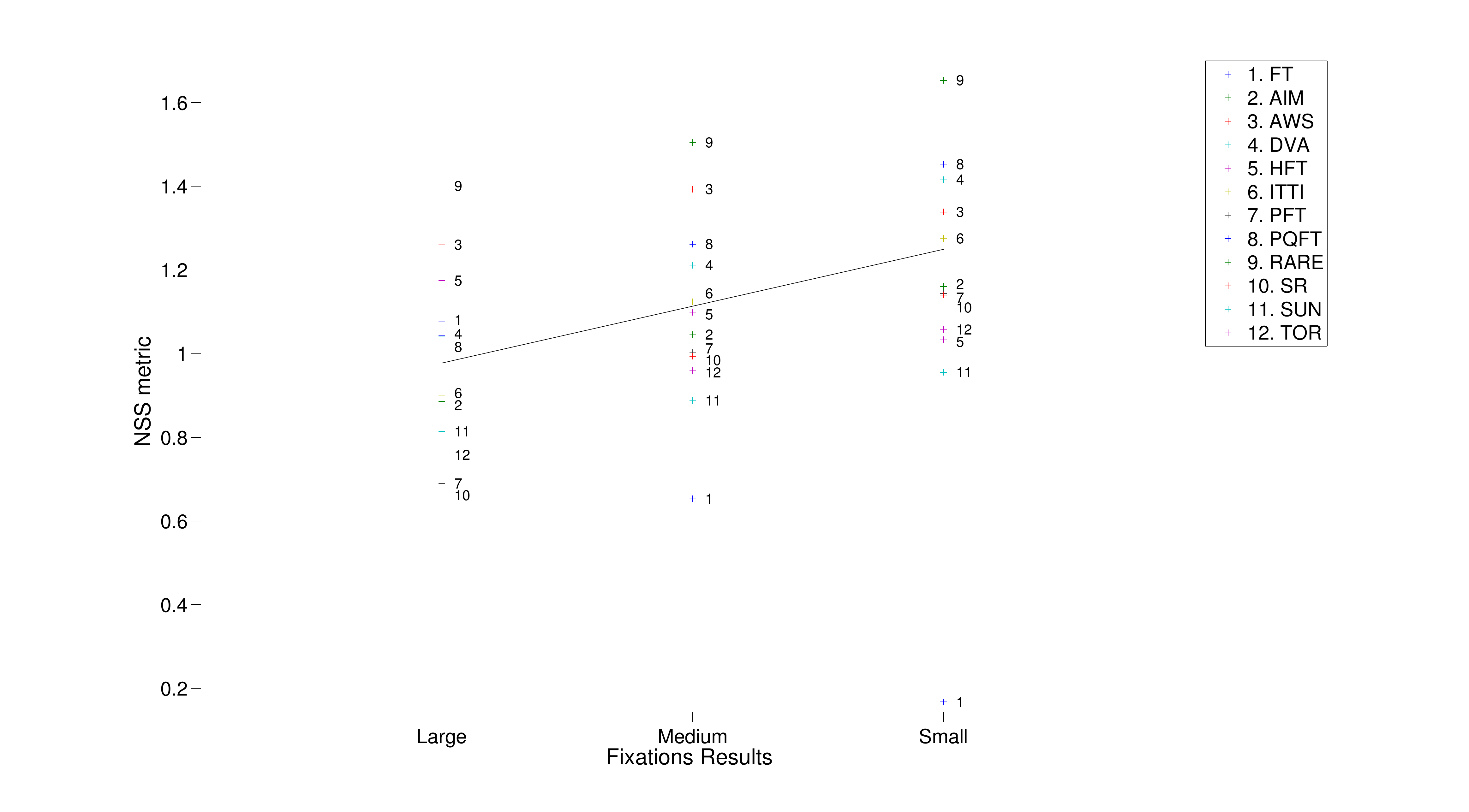}
 \caption{Eye fixations results on large, medium and small regions for AUCROC (top graph) and NSS (bottom graph)}
 \label{fig:exp21}
\end{figure}

\begin{figure}[h!]
 \centering
 \includegraphics[scale=0.28]{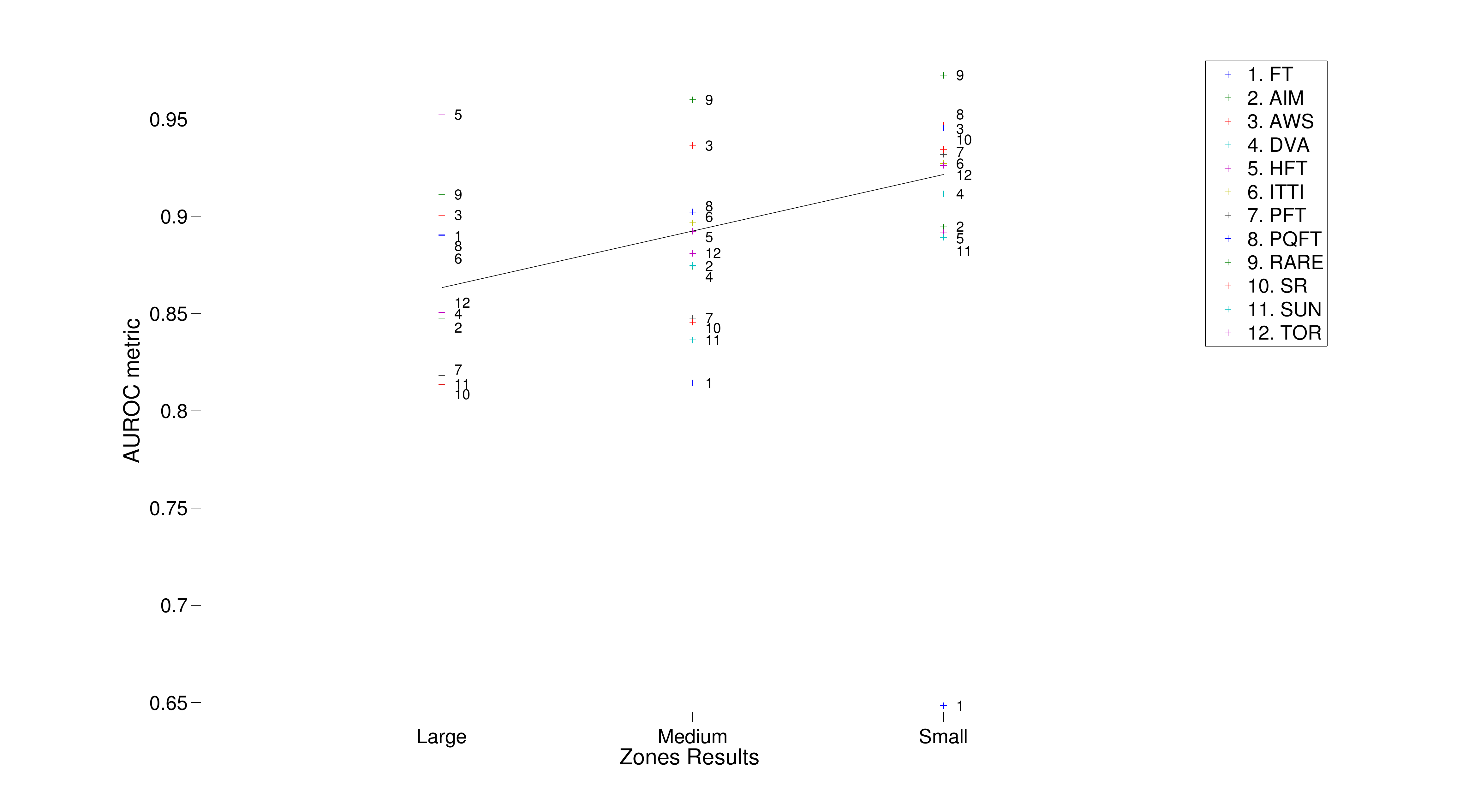}
  \includegraphics[scale=0.28]{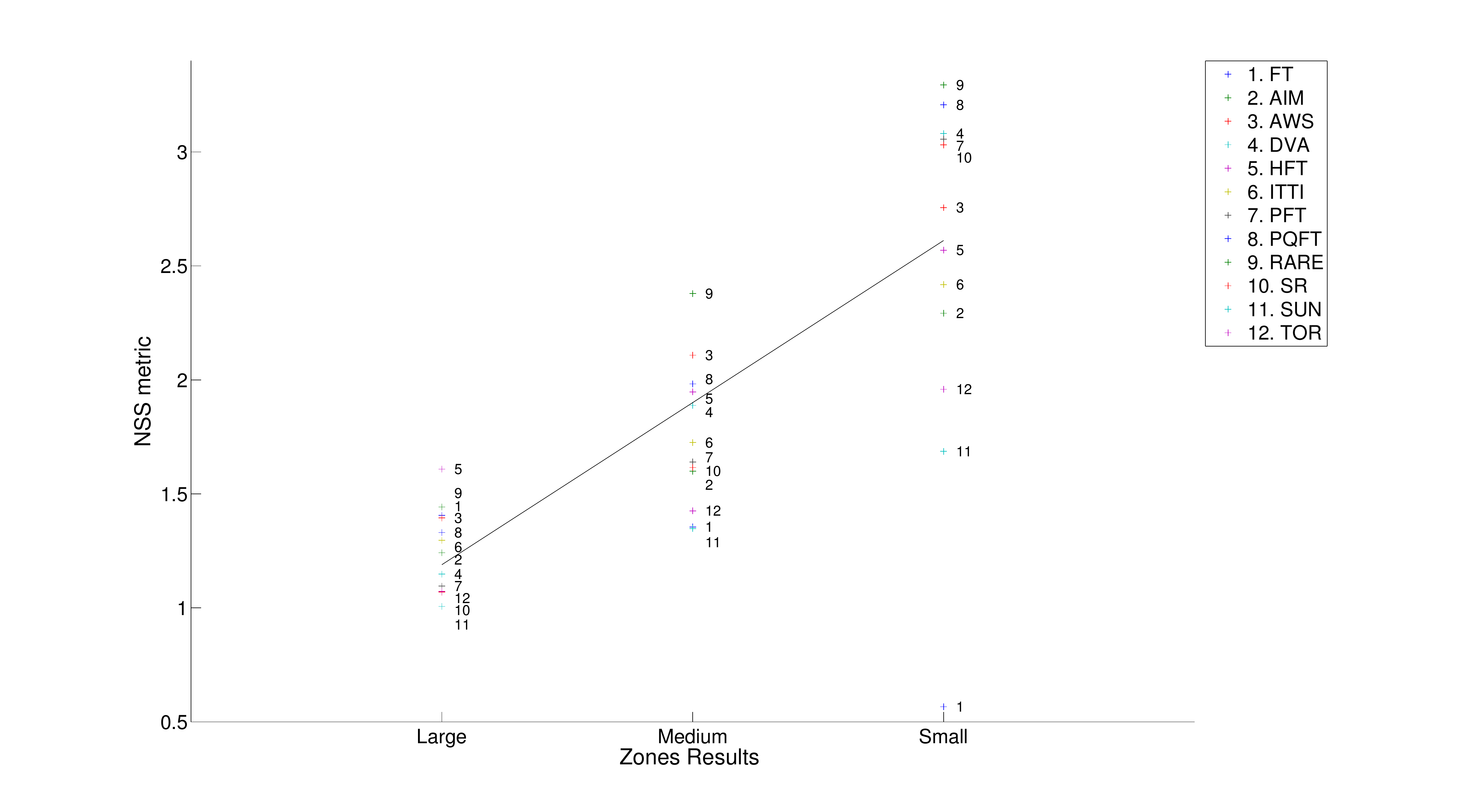}
 \caption{Labelled regions results on large, medium and small regions for AUCROC (top graph) and NSS (bottom graph)}
 \label{fig:exp22}
\end{figure}

Fig. \ref{fig:exp21} shows AUROC and NSS metrics into the three categories for eye fixations ground-truth and Fig. \ref{fig:exp22} for labelled regions ground-truth.
The mean tendency can be computed by a linear regression (black line on Fig. \ref{fig:exp21} and \ref{fig:exp22}). FT model was not used for regression building because it is clearly an outlier in all cases for medium and small salient regions. This model works well for large regions only.

Three general trends can be highlighted: 1) the small regions have higher scores than medium and large regions, 2) eye fixations results show less variation than labelled regions results and 3) the metrics exhibit different behaviors.  

The first observation is correct for almost all models. However, like FT, AWS is different: its score is better with large salient regions and its ranking decrease on small regions. On the other hand, PQFT model significantly increases (in terms of scores rank) for small regions. Compared to the other models FT and AWS are less good for small objects while PQFT is less good for large objects.
The second observation could be explained by the nature of the ground-truths. Indeed, the size of the segmented regions varies widely from one category to another. Eye fixations ground-truth is less affected by the size. For example, in Fig. \ref{fig:base}, the panda has a huge human label region compared to other labels from medium and small regions. The dispersion of human gaze between the panda and the other images is less important. 
Finally, the third observation is due to metrics definition: NSS depends more on the size of the salient regions than AUCROC which is more concerned about the regions of interest locations.\\

\begin{table}[htbp]
  \centering
  \caption{Concordance based on Friedman test and Kendall Coefficient for Large, Medium and Small regions}

\begin{tabular}{|c|c|c|c|}
  \hline
                  & \multicolumn{2}{c|}{Friedman test} & Kendall's concordance \\
                  & p-value  & $\chi^{2}$ (2, N=12) & W \\
  \hline
 AUROC Eye fixations & $6*10^{-2}$  &  5.64  & 0.84 \\
 NSS Eye fixations       & $1*10^{-3}$   &  13.27  & 0.85 \\
 AUROC Labelled regions      & $6*10^{-4}$  &  14.73  & 0.74 \\
 NSS Labelled regions            & $2*10^{-5}$  &  22  & 0.79 \\
  \hline
\end{tabular}
  \label{conc1}
\end{table}

To assess the coherence between categories, the same ranking-based statistical test is required than in experiment 1, but here on the means. We use the means because the number of images are different by categories. Therefore, the same Kendall's W coefficient than in experiment 1 shows us a smaller concordance. As shown in Tab. \ref{conc1}, the p-value is significant except for AUROC eye fixations (Bonferoni significant  threshold: 0.012). 
 It means that the ranks between models are statistically different between the three categories but the size of this difference in terms of ranking is relatively small. Indeed, the Kendall's concordance show a relatively strong agreement. Moreover, it is generally lower with labelled regions. The ranking between models is less well respected with labelled regions than with eye fixations ground-truth when going to category level. Here we may conclude that eye fixations ground truth is a slightly better (more stable) ground truth compared to the labelled regions especially for large salient regions.
In this experiment, the ranking are globally coherent (but less than between the two ground-truths). So, the size of the salient region can have a stronger impact on our assessment than the chosen ground-truth. The readers can see the behaviour of each model on Fig. \ref{fig:exp21} and \ref{fig:exp22}.


\subsection{Experiment 3: Effect of the chosen metrics }

\begin{figure}[h!]
 \centering
 \includegraphics[scale=0.25]{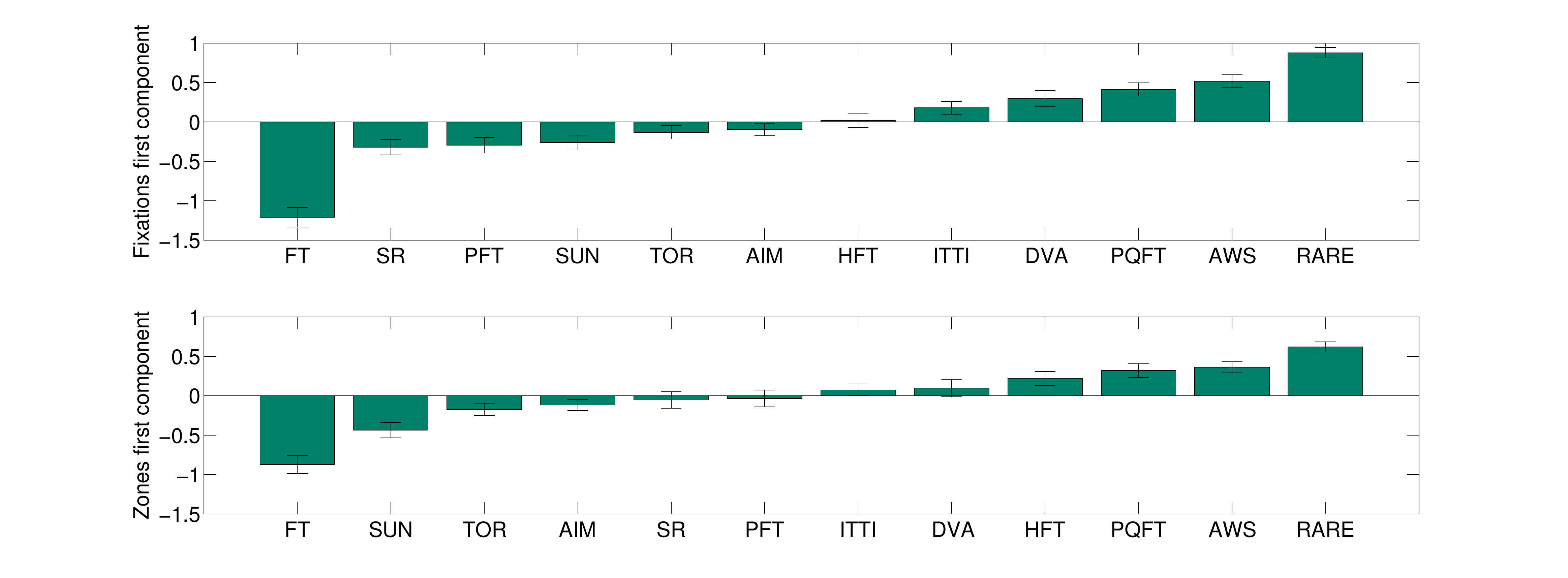}
 \caption{1st PCA component of metrics on LI database}
 \label{fig:zonesre1}
\end{figure}

In this experiment, the NSS and AUROC metrics were merged by PCA analysis to reduce the dimensionality of metrics. The first PCA-based component of each ground-truth were compared in terms of rank models
with the initial metrics. Fig. \ref{fig:zonesre1} shows results the first principal component analysis. The explained variances in Tab. \ref{conc2} are high and show us that the first components should be enough to explain the merge of the initial metrics. However, the zones ground truth (84\%) will be less representative that eye-fixations (91\%).  

The three top models are the same with eye fixations and labelled regions. Compared to the initial metrics (Fig. \ref{fig:fixre} and \ref{fig:zone}), this seems coherent. Moreover, the first component represents a large percentage of original variance of data.

An ANOVA test is then computed. The results were adjusted by linear regression to eliminate the effect of image salient regions covariance into the use of ANOVA.
They are similar in Fig. \ref{fig:zonesre1} but with smaller differences. This means that the effect of categories does not change the rank between models but only the values. 
Post-hoc tests were performed using a Tukey-HSD contrast \cite{univarie}. They show that the RARE model statically outperforms all the other models except AWS.

\begin{table}[htbp]
  \centering
  \caption{Explained variance which well represent the data and partial $\eta^{2}$ computed for PCA fixations and zones}

\begin{tabular}{|c|c|c|}
  \hline
                  & explained variance (\%) & partial  $\eta^{2}$ \\
  \hline
 PCA fixations & 91.43  & 0.316  \\
 PCA zones     & 84.37  & 0.168   \\
  \hline
\end{tabular}
  \label{conc2}
\end{table}

\section{Conclusion}
\label{sec:conclusion}

In this paper, we propose a new study of parameters affecting visual saliency assessment for real-life images. To achieve this goal, three experiments investigate basic questions to fairly evaluate saliency maps with human gazes or labelled regions.\\

First, a statistical ranking-based test shows that there are significant differences between eye fixations and manually segmented salient regions results but the concordance between the ranking of models is strong. Overall, the two ground truth methods provide coherent and interesting results.

In a second experiment, the properties of the salient regions (large, medium and small) are addressed with different degrees of accuracy by the saliency models. For most of them, small salient regions are better detected than medium and large salient regions. Moreover, the eye fixation ground truth seems more stable than the region labelled one mainly on the category of large salient regions. 

Finally, in a third experiment, a PCA analysis of metrics is used to newly assess saliency models. This gives coherent and more stable results. This merger let us provide only one ranking per ground truth which includes discriminant information from both AUCROC and NSS metrics. \\

\section*{Acknowledgments} 

N. Riche is supported by a FNRS/FRIA grant. 
M. Duvinage is a FNRS (Fonds National de la Recherche Scientifique) Research Fellow and the corresponding author to statistical technical considerations. 
M. Mancas is supported by Walloon region funding (Belgium).

This paper presents research results of the Belgian Network DYSCO (Dynamical Systems, Control, and Optimization), funded by the Interuniversity Attraction Poles Programme, initiated by the Belgian State, Science Policy Office. The scientific responsibility rests with its author(s).

\end{document}